\documentclass{article}

    \usepackage[nonatbib, final]{neurips_2019}

\usepackage[utf8]{inputenc} 
\DeclareUnicodeCharacter{2212}{-}
\usepackage[T1]{fontenc}    
\usepackage{hyperref}       
\usepackage{url}            
\usepackage{booktabs}       
\usepackage{amsfonts}       
\usepackage{nicefrac}       
\usepackage{microtype}      
\usepackage{graphicx}

\title{FireNet: Real-time Segmentation of Fire Perimeter from Aerial Video}

\author{
    Jigar Doshi\\
    CrowdAI\\
    \texttt{jigar@crowdai.com} \\
   \And
    Dominic Garcia \\
  Joint Artificial Intelligence Center \\
  \texttt{dominic.a.garcia17.mil@mail.mil} \\
  \And
  Cliff Massey \\
  CrowdAI \\
  \texttt{cliff@crowdai.com} \\
    \And
   Pablo Llueca \\
  CrowdAI \\
  \texttt{pau@crowdai.com} \\
   \And
   Nicolas Borensztein  \\
  CrowdAI \\
  \texttt{nic@crowdai.com} \\
   \And
     Michael Baird \\
  California Air National Guard \\
  \texttt{michael.d.baird.mil@mail.mil}
  \And
  Matthew Cook \\
  Joint Artificial Intelligence Center \\
  \texttt{matthew.b.cook18.mil@mail.mil} \\
  \And
     Devaki Raj \\
  CrowdAI \\
  \texttt{devaki@crowdai.com} 
}

\begin{document}

\maketitle

\begin{abstract}
  In this paper, we share our approach to real-time segmentation of fire perimeter from aerial full-motion infrared video. We start by describing the problem from a humanitarian aid and disaster response perspective. Specifically, we explain the importance of the problem, how it is currently resolved, and how our machine learning approach improves it. To test our models we annotate a large-scale dataset of $400,000$ frames with guidance from domain experts. Finally, we share our approach currently deployed in production with inference speed of $20$ frames per second and an accuracy of $92$ (F1 Score). 
\end{abstract}

\section{Introduction}

The effects of climate change continue to be felt around the globe. As the global climate is a complex and interconnected system, more and more of these effects are becoming evident that may not have been noticed before, even in the last few years. One particular impact of climate change that is clear, however, is the increase in wildfires—both in terms of frequency and intensity. Previously, wildfires were a perennial issue only in certain fire-prone locations, such as parts of Australia and the western U.S. Now, however, new geographies and biomes are dealing with wildfire due primarily to the drying of forests, which makes them more susceptible to fire \cite{ClimateAssessment}. Climate change is not the only variable, however, as human activity is also responsible for the increased spread of wildfire. As more and more homes and infrastructure are built along forest boundaries—or even within them—additional fuel is available for wildfires once they occur.

With increased intensity of wildfire comes increased intensity of the level of damage they do to the natural and built environment. In California alone, the 310 separate incidents in the 2018 wildfire season resulted in over 1.6 million acres burned, 93 fatalities, and more than 23,000 burned structures \cite{CalFire}. Damage to insured structures just in November 2018 was estimated to top $\$12$ billion, according to the California Insurance Commissioner’s office \cite{CalDamage1}. The monetary costs of wildfire have an enormous impact on a region, as property insurance companies, reinsurance companies, and individual policyholders must all attempt to cover losses. This has made it more difficult for those returning and rebuilding to find property insurance again, with a six percent increase in cases where the insurer did not renew property insurance from 2017 to 2018 \cite{CalDamage2, gupta2019creating, ioffe2015batch}.

Environmental impacts, too, are of concern after a wildfire. By burning hundreds or sometimes thousands of acres of vegetation, wildfires release the carbon trapped within that vegetation into the air, mainly as carbon dioxide. Those same 2018 wildfires in California released an estimated 68 million metric tons of carbon dioxide, roughly the same as the state’s annual emissions from electricity generation \cite{CalDamage3}. This by-product of the fire reinforces the warming effects of climate change, exacerbating the underlying problem.

Even with potential carbon mitigation and active forestry management efforts in the near- and mid-term, we can expect the upward trend in wildfire frequency, intensity, and damage to continue. With this in mind, it is clear that firefighters and other first responders will need improved or new tools to better understand the extent and behavior of wildfire so that they may plan their response accordingly. In California, the Department of Forestry and Fire Protection (CAL FIRE) is the agency responsible for the stewardship and protection of the roughly 31 million acres of the state’s wildland.

The current procedure for CAL FIRE is to request aerial data to identify and monitor the extent of an active fire is a multi-step process that enables the agency to make a request through government channels for assistance from the California Air National Guard (CA ANG). Once a CA ANG aerial asset and sensor is within range of the fire, the CA ANG team uses near-real-time infrared (IR) full-motion video (FMV) feed to locate the fire perimeter, placing the sensor’s reticle on and around the perimeter of the fire. A team of intelligence analysts, including all source analysts, imagery analysts, and incident awareness and assessment coordinators assess the FMV frames looking for the fire perimeter. The sensor point of interest (where the sensor is looking on the ground) is synced with a mapping tool on a 2D geospatial mapping platform. From here, an analyst must manually point-and-click for hours on-end to create a fire perimeter polygon, often while working around-the-clock shifts. The fire perimeter analysts will analyze the sensor crosshairs, intensity of the IR returns, digital terrain elevation data, and geographical and topographical features, all while looking for possible false positives (such as roads, bodies of water, and flares) in order to make an assessment of where they think the fire perimeter is currently located. Depending on the size and extent of the fire, this process can take hours to complete—even for just a single section of fire perimeter.

This is where automation—specifically in the form of machine learning—can provide a force multiplier, as human/machine teaming reduces the cognitive burden on analysts, freeing them up to focus on higher-order tasks—like contextual assessments, sensemaking, and decision-making—that machines currently cannot accomplish. Because speed and accuracy are both critical to response efforts, machine learning algorithms and a more automated process can be used to provide first responders with more accurate information about the fire perimeter more often and with higher accuracy. Wildfires can spread at rates of 7-10 miles per hour in forested regions, and even more quickly in grasslands. Near-real time updates of the fire perimeter are therefore needed to inform the public and to coordinate response efforts. Accuracy, too, is important, especially as it relates to the precise geolocation of the fire perimeter. Not only will automating the fire perimeter speed up the analysis process, it will also mean that high demand, low density FMV assets can cover more ground. By taking the human out of the loop and putting them on the loop we anticipate a return on investment in both human capital as well as sensing resources in flight resulting in overall better situational awareness

\section{Approach}

We cast this fire perimeter detection as a segmentation task. Our task is to compute a binary mask of the fire for every input frame of the video. For video segmentation applications, in order to achieve temporal consistency,  LSTM- or RNN-like layers are used. However, they are too computationally expensive for an on-device, real-time segmentation application. We build our model in two steps. First, we train a standard over-parameterized model to achieve the best performance. Following that, we prune our models to trade off accuracy and inference speed. 

One of the key issues for video segmentation over naive frame-by-frame segmentation is temporal consistency, which causes poor overall performance \cite{hairSegmentation}. To address these issues, we explore the following two strategies: 1) Adding more frames for context using 3D Convs and 2) Adding the previous prediction as additional channels.

The second goal is to make these models lightweight enough to run above 15 frames per second on a standard K80 Nvidia GPU. This is a hard requirement given that these models are built for real-time segmentation applications. In order to achieve this inference speed, we prune our models along these three dimensions: 1) Input Resolution, 2) Model Thickness, and 3) Model Depth

A typical pruning algorithm in the literature \cite{net-compression-suau, net-compression-hu} is a three-stage pipeline, i.e. training (a large model), pruning, and fine-tuning. During pruning, according to a certain criterion, redundant weights are pruned and important weights are kept to best preserve the accuracy. However we found in our experiments that fine-tuning a pruned model only gives comparable or worse performance than training that model with randomly initialized weights. This observation was also validated on wide range of datasets by \cite{liu2018rethinking}. Hence all our pruning experiments do not follow this three stage approach; we always train the pruned model from scratch. 

\subsection{Dataset}

Our dataset consist of short videos (150 frames) captured over the past few years for various wildfires around the U.S. This massive dataset has roughly 400,000 frames annotated. Out of these 400K frames around 100K frames have an active fire in the video. All the videos are captured via on-board infrared (IR) sensor, since the fire perimeter would rarely be visible in the standard RGB spectrum due to smoke and other factors. 

The specific computer vision task is to segment out burning or burnt regions in the video. The technical term is fire perimeter, which means entire length of the outer edge of the fire \cite{FireTerms}. To annotate these videos we consulted subject matter experts from CAL FIRE and the California Air National Guard. Subsequently, once the definitions were agreed upon, we used our annotation vendors to annotate this dataset with constant QA from the experts.

\begin{figure}
  \centering
  \includegraphics[scale=0.35]{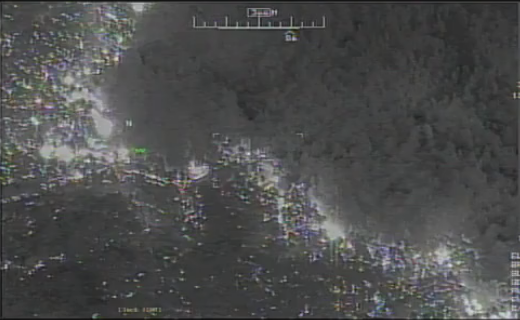}
  \includegraphics[scale=0.35]{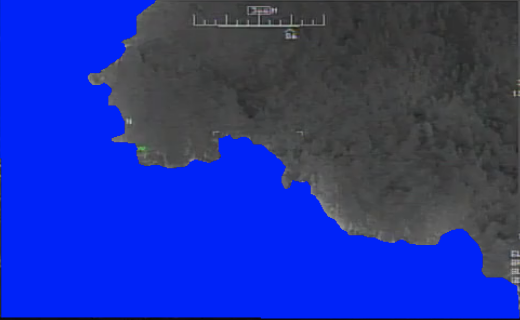}
  \includegraphics[scale=0.35]{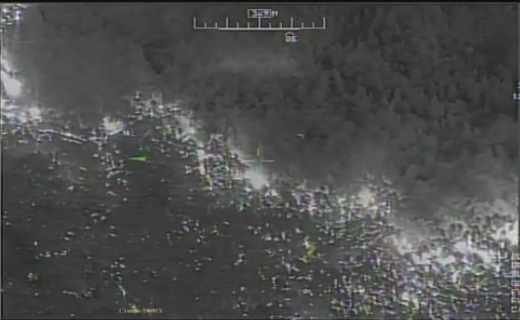}
  \includegraphics[scale=0.35]{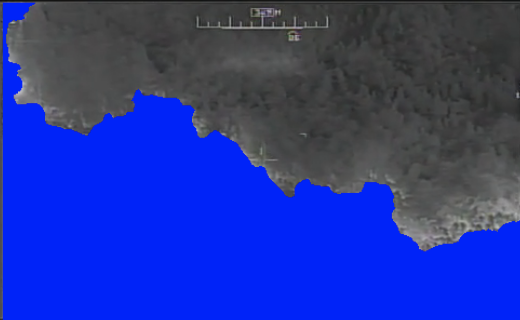}
  \caption{Left is input to the model and right is the ground truth annotation}
\end{figure}


\section{Network Architecture}
Our main model was inspired by the family of U-Net architectures \cite{Ronneberger2015-ia,doshi2018residual}, where low-level feature maps are combined with higher-level ones, which enables precise localization. This type of network architecture was designed to effectively solve image segmentation problems, particularly in the medical imaging field. This hour-glass model is generally a default choice for segmentation challenges in Kaggle.

The encoder of the model consists of resnet blocks \cite{He2015-cq} with the addition of batch norm \cite{Szegedy2015-bn} and a total of 8 downsampling layers. We decided to keep a constant number of 128 feature maps throughout the network; we will vary this parameter while pruning the model. The decoder is similar to the encoder, where instead of max-pooling, we use deconvolution layers to upsample with a skip connection from the corresponding encoder, combining deep representations of the prior decoder layer with more precise spatial representations from the corresponding encoder layer.  All weights are initialized with the He norm \cite{He2015-td} and all convs are followed by batch norm and the activation function. We used a leaky RELU with a slope of $-0.1x$ as our activation function.

The final head consists of a $1\times1$ convolution followed by a hard sigmoid activation function. We use this activation function instead of a standard sigmoid because of our choice of a sensitive dice loss function. In the case of the soft sigmoid function, the model activation only approaches the extreme values (0 and 1) values which leave a lot of small residual error that adds up and diffuses the model’s ability to focus on ‘harder’ error signals.

\subsection{3D UNet Model}

We take our existing UNet architecture and add an additional dimension of input. We add the previous 8 frames as input to the model. All the 2D Convs in the encoder and decoders were converted into 3D and the final head was appropriately reshaped to get an output for the latest frame. Surprisingly, this approach did not yield any significant performance improvement however the model was very slow. There is another logical option of both input and output as 8 frames at a time. However, due to the real-time requirements of the system, we cannot have any latency of this nature. Due to the above-mentioned reasons, we abort this line of model exploration.

\subsection{Adding Previous Frame Predictions (PrevPred)}

The idea is to add the previous predictions of the model back as input for the current frame’s prediction. This technique has shown to improve model consistency in video segmentation tasks[hair seg paper]. After some hyperparameter search, we found the best result when we added $t-1, t-3, t-5$ frames’ prediction into the model. Adding these frames slightly slows down the model. However, we see a consistent improvement in the performance as shown in table \ref{Results}. 

\subsection{Pruning}

We prune the model's depth from 8 to 4 without too much loss of performance. In addition, we pruned the model’s thickness down. In the original architecture, we had 128 channels and we prune it down to 64 for the first layer and 32 for the rest. In both the pruning mentioned above pruning any further caused a sharp dip in model performance. We also experimented in reducing the input resolution to the model however that always led to a significant drop in performance, so we kept it at the original resolution.

\section{Training Setup}
\subsection{Data Augmentation}

We apply a host of data augmentations to the training data during training. We augment the images fed into all the models by randomly scaling between 1/1.05 and 1, rotate, flip, salt and pepper noise, shearing between −5 and 5 degrees. Each image had a 10\% chance of having the above augmentations applied. Not surprisingly, adding various data augmentation improved the model performance. It especially improved the model’s temporal consistency. However, even stronger data augmentation reduced model generalization. We think this might be due to the over-perturbation of the training distribution away from the real distribution. 

When we add the previous predictions mask back into the model, we apply a few data augmentation techniques. 
\begin{itemize}
    \item \textbf{Empty Mask:} We train the model to work well when starting a new clip by adding blank masks for previous frames
    \item \textbf{Affine Transformation: }We perturb the masks with random transformations during training. Small perturbations train the network to be robust to noise, while large perturbation trains the model to ignore these masks and just use the pixels. 
    \item \textbf{Ground Truth Masks}: Early on during the training instead of using the model's predictions from previous frames we use the ground truth annotations. This approach helps the model to learn quickly and also leads to slightly better performance.

\end{itemize}

\subsection{Loss Function}
The dice similarity coefficient (DSC) measures the amount of agreement between the model prediction and the ground truth. It is a widely used metric in high class imbalance segmentation tasks \cite{Milletari2016-jm,Shen2018-id} 




We use a continuous version of the Dice score that allows differentiation and can be used as a loss function in optimization of our network using stochastic gradient based methods.

\begin{equation}
    \mathcal { L } _ { D S C } = - \frac { 2\sum _ { i } ^ { N } s _ { i } r _ { i } } { \sum _ { i } ^ { N } s _ { i } + \sum _ { i } ^ { N } r _ { i } + \epsilon}
\end{equation}

where $s_i$ represents a continuous value from the model for each pixel which is typically an output from a $sigmoid$ or $softmax$ activation function. $r_i$ represents the ground truth annotation for each pixel. $\epsilon$ is a smoothing factor typically set to $1.0$

We experimented with a baseline binary cross-entropy loss and found it to be consistently worse than dice loss across all the configurations.

\subsection{Optimization}
We trained all the models with ADAM \cite{Kingma2014-vd} using $\beta1$ as 0.9 and $\beta2$ as 0.999 without any weight decay. Unless stated otherwise, we started the training with a learning rate of $2e^{-4}$.  We reduce the learning rate by a factor of $2$ after observing no improvement in the validation loss for 10 epochs. All the models were trained for $150$ epochs.

\section{Results}
In Table \ref{Results}, we show our quantitative results. We measure the inference speed in terms of number of frames per sec ($fps$) forward propagated on a single gpu (Nvidia K80). Each column represents a model setup. First column (Basic UNet) is the baseline model which we would like to ideally match in accuracy while achieving a much higher inference speed. The second column is the model where we add previous predictions back into the model for subsequent predictions. Unsurprisingly, the inference speed slows down since the input dimension was increased. The third column is our best pruned model without the addition of previous predictions. This setup achieves the best inference speed however it suffers in terms of accuracy. Finally, the model we use in production is the fourth column model where we add those previous predictions back into the pruned model. This model gives us the right balance between inference speed and accuracy. 

\begin{table}
  \caption{Each column represents different model and their corresponding inference speed measured in frames per second (fps) and accuracy as measured in F1 score}
  \label{Results}
  \centering
  \begin{tabular}{llllll}
    \toprule
    Model     & Basic UNet     & UNet+ PrevPred & Pruned w/o PrevPred & Pruned + PrevPred\\
    \midrule
    Inference Speed & $5 fps$  & $3 fps$  & $22 fps$ & $20 fps$ \\
    F1 Accuracy     & $94$ & $95$  & $86$ & $92$ \\
  
    \bottomrule
  \end{tabular}
\end{table}

\section{Conclusion and Future Work}
In this work, we show one of the first large-scale approaches to real-time mapping of fire perimeter in disaster situations. We motivate it by sharing the current state of affairs and why automating it would be extremely beneficial during disaster situations. We share our current approach to the problem, which shows good performance both in terms of accuracy and inference speed. In the future, after necessary considerations, we hope to share this dataset and let the community benefit from it. 

\medskip
\bibliographystyle{unsrt}
\bibliography{neurips_2019}

\end{document}